\documentclass[letterpaper, 10 pt, conference]{ieeeconf}

\IEEEoverridecommandlockouts

\usepackage{algorithm}
\usepackage[noend]{algpseudocode}
\usepackage{tikz}
\usepackage{subfigure}
\usepackage{amsmath}
\usepackage{mathrsfs}
\usepackage{enumerate}
\usepackage{eqparbox}
\usepackage{latexsym}
\usepackage{amsfonts}
\usepackage{{todonotes}}
\usepackage{verbatim}

\usepackage{fainekos-macros}

\usepackage{ifthen}
\usepackage{color,soul}
\newboolean{HIGHCOMM}
\setboolean{HIGHCOMM}{false}
\newcommand \yhl[1]{\ifthenelse{\boolean{HIGHCOMM}}{\textcolor{blue}{#1}}{#1}}
\newboolean{SHOW_WHAT_IT_WAS}
\setboolean{SHOW_WHAT_IT_WAS}{true}
\newcommand \shl[1]{\ifthenelse{\boolean{SHOW_WHAT_IT_WAS}}{\yhl{\st{#1}}}{}}
\newboolean{TECHREPORT}
\setboolean{TECHREPORT}{true}

\title{\LARGE \bf
DisCoF$^+$: Asynchronous DisCoF with Flexible Decoupling for Cooperative Pathfinding in Distributed Systems 
}

\author{
Kangjin Kim, Joe Campbell, William Duong, Yu Zhang and Georgios Fainekos
\thanks{This work has been partially supported by award NSF CNS 1116136, the ARO grant W911NF-13-1-0023, the ONR grants N00014-13-1- 0176, N00014-13-1-0519 and N00014-15-1-2027.}%
\thanks{K. Kim, J. Campbell, W. Duong, Y. Zhang and G. Fainekos are with the School of Computing, Informatics and Decision Systems Engineering, Arizona State University, Tempe, AZ 85281, USA {\tt\small \{Kangjin.Kim, jacampb1, tbduong, Yu.Zhang.442, fainekos\}@asu.edu} }
        }%

\begin{document}

\maketitle

\newcommand{\planner}{P}

\newtheorem{theorem}{Theorem}[section]
\newtheorem{corollary}[theorem]{Corollary}
\newtheorem{lemma}[theorem]{Lemma}
\newtheorem{claim}[theorem]{Claim}
\newtheorem{proposition}[theorem]{Proposition}

\newtheorem{definition}{Definition}[section]

\newcommand{\qed}{\hfill \ensuremath{\Box}}

\begin{abstract}

In our prior work, we outlined an approach, named DisCoF, for cooperative pathfinding in distributed systems
with limited sensing and communication range. 
Contrasting to prior works on cooperative pathfinding with completeness guarantees, 
which often assume the access to global information,
DisCoF does not make this assumption.
The implication is that at any given time in DisCoF, the robots may not all be aware of each other, which is often the case in distributed systems.
As a result, DisCoF represents an inherently online approach since coordination can only be realized in an opportunistic manner between robots that are within each other's sensing and communication range.
However, there are a few assumptions made in DisCoF to facilitate a formal analysis, 
which must be removed to work with distributed multi-robot platforms.
In this paper, we present DisCoF$^+$, which extends DisCoF by enabling an asynchronous solution, 
as well as providing flexible decoupling between robots for performance improvement.
We also extend the formal results of DisCoF to DisCoF$^+$.
Furthermore, we evaluate our implementation of DisCoF$^+$ and demonstrate a simulation of it running in a distributed multi-robot environment.
Finally, we compare DisCoF$^+$ with DisCoF in terms of plan quality and planning performance.
\end{abstract}



\section{INTRODUCTION}
\label{sec:introduction}

While cooperative pathfinding in multi-robot systems has many applications, 
it is also fundamentally hard to solve (i.e., PSPACE-hard \cite{hopcroft1984}).  
The difficulty lies in the potential {\em coupling} between robots:
when robots are completely decoupled 
(e.g., when robots do not impose constraints on each other's plan to the goal), 
cooperative pathfinding becomes polynomial-time solvable.\footnote{
A single robot pathfinding problem is polynomial-time solvable.
}
As a result, most recent approaches (e.g., \cite{Sharon201540, Silver05aiide, standley2010,standley2011})
for pathfinding concentrate on how to identify the dependencies between robots, 
in order to couple robots only when necessary to achieve computational efficiency for many problem instances.

In these approaches, the solution is constructed for a subset of robots (i.e., robots that must be coupled) at any time,
which are assumed to be decoupled with the remaining robots.
The computational complexity is exponential only in the maximum number of robots in these subsets. 
While optimistic decoupling can lose optimality and even completeness (e.g., \cite{Silver05aiide}),
pessimistic decoupling can only handle situations in which robots are loosely coupled (e.g., \cite{standley2010}).

Meanwhile, to ensure completeness, these approaches often assume access to global information, which includes knowledge about the current positions of the robots, and the robots' individual plans to their respective goals.
With this information, any robot can consider all other robots when creating its own plan.   
While this assumption can be made in many common applications of cooperative pathfinding where planning can be centralized and performed offline
(e.g., cooperative pathfinding in computer games),
it does not hold in distributed systems due to limited sensing and communication range.

In our prior work \cite{yu-dars-2014}, we introduced a window-based approach, called DisCoF, 
for cooperative pathfinding in distributed systems with limited sensing and communication range.
In DisCoF, the window size corresponds to the sensing range of the robots.
Robots can communicate with each other either directly or indirectly.
If they are within sensing range, then robots may communicate directly.
However, if the robots are out of sensing range it is still possible to communicate indirectly through other robots using a communication relay protocol.
This allows for coordination beyond a single robot's sensor range.

To ensure completeness, 
DisCoF uses a flexible approach to decoupling robots such that they can transition from optimistic to pessimistic decoupling 
when necessary. 
Robots are assumed to be fully decoupled initially.
During the online pathfinding process, 
robots only couple together when necessary (i.e., when there are {\em predictable conflicts} \cite{yu-dars-2014}).
Since access to global information is not assumed, the creation of local couplings (i.e., subsets of robots) may not be sufficient due to the danger of live-locks.
In such cases, a mechanism (called {\em push and pull}) is introduced in which robots in a local coupling can form a {\it coupling group} \cite{yu-dars-2014} in order to coordinate more closely.
Robots in a coupling group move to their goals sequentially in a certain order while keeping others (i.e., those that have not yet reached their goals) within communication range.
Coupling groups may increase in size (e.g., when previously undetected robots come within sensing range of a robot in the coupling group) and decrease in size (e.g., when robots reach their goals).
This mechanism can potentially lead to a global coupling.


\textbf{Contributions:}
In this paper, we introduce an asynchronous variant of DisCoF, refereed to as DisCoF$^+$, in order to remove DisCoF's required assumption that time steps are synchronized.
\yhl{The Major difference of DisCoF$^+$ from DisCoF is that this one runs on the individual robot, but the previous one runs on a group of robots, depending on the synchronized time step for the entire robots. In order to make it work, we provide an asynchronous algorithm with its communication strategy.}
Then, we introduce a new decoupling strategy in DisCoF$^+$ that allows robots to transition between optimistic and pessimistic decoupling with the goal of improving efficiency.
Furthermore, we demonstrate a simulation of DisCoF$^+$ in a distributed multirobot environment modelled in Webots in addition to providing the results of numerical experiments with which the performance of DisCoF and DisCoF$^+$ are compared in terms of computation time and length of plans.


\section{RELATED WORK}
\label{sec:related-work}

To address the cooperative pathfinding problem, 
researchers have used a compilation approach \cite{LiuS13ijrr,AyanianRK12wdecns,DesarajuH12ar,YuL12wafr},
in which the problem is first transformed into other related problems,
and then the existing solutions or algorithms for these problems can be applied.
Abstraction methods to reduce the search space have also been used \cite{sturtevant2006,ryan2007}.
However, due to the inherent complexity of the problem, these approaches are unscalable.
While approaches that constrain the topologies of the environment \cite{wang08,Parker09,peasgood2008} 
can significantly reduce the complexity,
they cannot be applied to general problem instances.

Given that pathfinding for a single robot is polynomial-time solvable, 
it is clear that the complexity is a result of coupling between robots.
As a result, researchers have concentrated on various ways to decouple robots.
For approaches that perform optimistic decoupling,
robots are considered as coupled only when necessary. 
One of the representative approaches is hierarchical cooperative A$^*$ (HCA$^*$ \cite{Silver05aiide}), in which robots plan one at a time while respecting plans that have already been calculated.
To limit the influence of the previous robots on the following robots, 
a windowed HCA$^*$ is introduced to restrict this influence based on a pre-specified window size \cite{Silver05aiide}. 
Recently, an extension of WHCA$^*$ (CO-WHCA$^*$ \cite{bnaya}) 
is introduced to further reduce this influence based on the notion of {\em conflicts}.
Although many problem instances can be solved efficiently, 
optimistic decoupling in these approaches leads to the loss of optimality and completeness.

One of the earlier approaches that performs decoupling while maintaining optimality and completeness
relies on pessimistic decoupling \cite{standley2010},
which couples robots when any conflicts are detected in their entire individual plans.
As a result, this approach tends to over-couple and hence remains intractable for many problem instances.
More recent approaches relax optimality to achieve better efficiency \cite{LunaB11iros, WildeMW13aamas, standley2011}.   
However, to maintain completeness, these approaches assume access to global information and therefore are inapplicable to distributed systems in which robots have limited sensing and communication range.

While there are extensible approaches to distributed systems (e.g., \cite{jansen2008})
and approaches that are designed for distributed systems (e.g., \cite{frazzoli,clark2003}), 
they do not provide completeness guarantees.
The difficulty lies in planning without access to global information, which is addressed in \cite{yu-dars-2014}.



\section{DisCoF}
\label{sec:approach}

In this section, we provide the problem formulation and review DisCoF \cite{yu-dars-2014}.
Extensions to DisCoF (i.e., DisCoF$^+$) are discussed in Section \ref{sec:tech}.

\subsection{Problem Formulation}

Given an undirected graph $G(V, E)$, and a set of robots $\mathcal{R}$, 
the initial locations of the robots are $\mathcal{I}\subseteq V$, and the goals are $\mathcal{G}\subseteq V$. 
Any robot can move to any adjacent vertex in one time step or remain where they are. 
A plan $\mathcal{P}$ is a set of individual plans of robots, 
and $\mathcal{P}[i]$ denotes the individual plan for robot $i \in \mathcal{R}$. 
Each individual plan is composed of a sequence of actions.
For simplicity, in this paper each action is represented by the next vertex to be visited.
For example, $\mathcal{P}_k[i]$ $(k \geq 1)$ denotes the action to be taken at time step $k - 1$ (or the vertex to be visited at $k$) for robot $i$.
$\mathcal{P}_{k,l}[i]$ $(k\leq l)$ denotes the subplan that contains the actions from $\mathcal{P}_k[i]$ to $\mathcal{P}_l[i]$.
The goal of cooperative pathfinding is to find a plan $\mathcal{P}$, 
such that robots start in $\mathcal{I}$ and end in $\mathcal{G}$ without any {\em conflicts} (defined below). 
The location of robots at time step $k$ is denoted by $\mathcal{S}_k$, 
and the location of robots after executing plan $\mathcal{P}$ from location $\mathcal{S}$ is denoted by $\mathcal{S}(\mathcal{P})$.
Hence, $\mathcal{S}_0 = \mathcal{I}$, $\mathcal{S}_0(\mathcal{P})= \mathcal{G}$, and $\mathcal{S}_k =  \mathcal{S}_0(\mathcal{P}_{1,k})$.
A {\em conflict} happens at time step $k$, if the following is satisfied:

\begin{equation}
\mathcal{S}_{k}[i] = \mathcal{S}_{k}[j] \lor (\mathcal{S}_{k}[i] = \mathcal{S}_{k - 1}[j] \land \mathcal{S}_{k - 1}[i] = \mathcal{S}_{k}[j])
\label{eq:conf}
\end{equation}
in which $i \in \mathcal{R}$, $j \in \mathcal{R}$ and $i \neq j$. 
\yhl{If two robots move to the same place at time $k$, the first condition holds. If two robots switch their locations in two consecutive time steps from $k-1$ to $k$, the second condition holds.}
\ifthenelse {\boolean{TECHREPORT}}
{\yhl{Figure \ref{fig:mr:simple} shows the first condition.}}
{
}

Each robot has a planner that can compute a shortest path, $\planner(u, v)$, that moves a robot from vertex u to v.
The length of $\planner(u, v)$ is denoted as $\mathcal{C}(u, v)$, i.e., $\mathcal{C}(u, v) = |\planner(u,v)|$.
The following simplifying assumptions are also made in DisCoF:

\begin{enumerate}
	\item Robots are homogeneous and equipped with a communication protocol for message relay. 
	\item Robots know $G$ and are synchronized at every time step.
\end{enumerate}

Initially, for each robot $i$, the individual plan is constructed as $\mathcal{P}[i] = \planner(\mathcal{I}[i], \mathcal{G}[i])$.
Robots then start executing their individual plans until conflicts can be predicted 
(i.e., {\it predictable conflicts} in \cite{yu-dars-2014}) at time step $k$.
In such cases, the individual plans of robots that are involved are updated from $\mathcal{P}_{k + 1}$ to avoid these conflicts.

\subsection{Optimistic Decoupling}

\begin{figure}
\centering
\includegraphics[width=6cm]{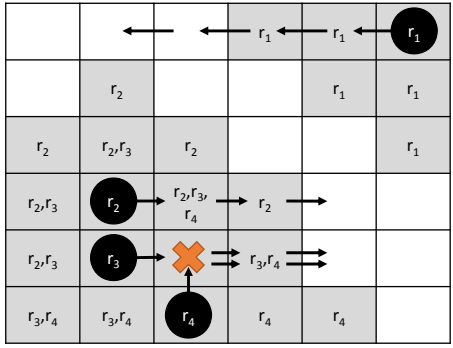}
\caption{\cite{yu-dars-2014}: Scenario that illustrates OC and IC.
Two OCs are present $\{r_1\}$ and $\{r_2, r_3, r_4\}$, out of which one has an IC $\{r_3,r_4\}$ with a predictable conflict. 
The sensing ranges of the robots are shown in gray. 
The arrows show the next few steps in the individual plans.}
\label{fig:mr:simple}
\end{figure}

In DisCoF, the window size corresponds to the sensing range of the robot. 
To reduce communication, a robot is allowed to communicate with other robots when it can sense them. 
However, robots that cannot sense each other can communicate using the message relay protocol through other robots.
A closure of the set of robots that can communicate (directly or via message relay) in order to coordinate is called an {\em outer closure} (OC).
In an OC, there can be multiple predictable conflicts. 
A closure that contains agents with potential conflicts is the {\em inner closure} (IC) of the OC.
Figure \ref{fig:mr:simple} shows an example of OC and IC.
For details, refer to \cite{yu-dars-2014}.

In DisCoF, decoupling is optimistic initially, and gradually becomes more pessimistic during the online planing process when necessary.
Given an OC with predictable conflicts, 
in {\em optimistic decoupling} 
DisCoF updates the individual plans of robots to proactively resolve these conflicts,
while avoiding introducing new conflicts within a finite horizon
(which is specified by a parameter in DisCoF).
The finite horizon is key to efficiency since the resolution for conflicts in the far future is likely to waste computation efforts given the incomplete information (e.g., other robots in the environment).
Note that the window size (i.e., sensing range) in DisCoF represents a horizon for detecting conflicts.

To ensure that robots are jointly making progress towards their goals, 
DisCoF uses the notion of {\em contribution value}.
In order to resolve conflicts, plans are updated in a process known as conflict resolution.
In this process, each robot is associated with a contribution value when using optimistic decoupling.
If this process is successful,
robots continue as fully decoupled.
The contribution value is also used to determine cases when optimistic decoupling is insufficient,
in which the resolution process would fail due to potential live-locks.
When there are no potential live-locks, 
it is shown that optimistic decoupling is sufficient for robots to converge to their goals. 
Otherwise, robots within the OC use the following \textit{pessimistic decoupling} process.

\subsection{Pessimistic Decoupling}
\label{subsec:pess}

In DisCoF, when there are potential live-locks,
robots within an OC transition to {\em pessimistic decoupling} by remaining within each other's communication range (whether direct or indirect).
These robots are referred to as a {\em coupling group}, 
and this coupling group executes a process known as push and pull, which allows it to merge with other groups and robots.
Thus, the level of coupling gradually increases.
In this way, DisCoF can naturally transition robots to be fully coupled when necessary.

In push and pull, 
robots move to goals one at a time according to the priorities of subproblems (first introduced in \cite{WildeMW13aamas}). 
However, due to the incompleteness of information in distributed systems, 
the priorities will not be fully known.
As a result, DisCoF employs the following process.
At time step $k$, for each coupling group that has been formed, DisCoF will:

\begin{enumerate}
	\item Maintain robots in the group within each other's communication range;
	\item Move robots to goals one at a time based on a relaxed version of the priority ordering, 
which is consistent to that in \cite{WildeMW13aamas}; 
	\item Add other robots or merge with other groups that introduce potential conflicts with robots in the current group as they move to their goals.
\end{enumerate}

Unless there are potential conflicts, each coupling group progresses independently of other robots and coupling groups.
\ifthenelse {\boolean{TECHREPORT}}
{
\yhl{These processes are} described in Alg. \ref{alg:coord}:



\begin{algorithm}
\caption{Pessimistic Decoupling in DisCoF for a coupling group $\omega$, given the current time step $k$, the environment $G$, current locations $\mathcal{S}$ and goal locations $\mathcal{G}$}
\begin{algorithmic}[1]
\State $r$ $\gets$ $\bot$ \Comment Initialize the leader robot $r$
\While {$\exists i \in \omega$ s.t. $\mathcal{S}_k[i] \neq \mathcal{G}[i]$} \label{alg:coord:termination_condition}
\State $\tupleof{\psi,\phi} \gets \text{\sc SenseConflict($P$, $\omega$, $\mathcal{S}$, $k$, $\mathcal{W}$)}$
\If {$\psi = \emptyset$}
\State $k \gets k + 1$	\Comment Increase the time step by 1
\State $G' \gets \tupleof{G, \omega, S, \mathcal{G}}$
\State $\tupleof{\mathcal{S}, P, \mathcal{W}} \gets \text{\sc ProceedOneStep}(G', P, k)$
\Else
\State $\omega \gets \omega \cup \psi$	\Comment Merge conflict robots with $\omega$
\State $\tupleof{f, D} \gets \text{\sc AssignAgentsToSubP($G$, $\omega$, $\mathcal{S}$, $\mathcal{G}$)}$
\State $H \gets \text{\sc computePriority($G$, $\omega$, $f$, $D$, $\mathcal{S}$, $\mathcal{G}$)}$
\State $r \gets \bot$
\EndIf
\If {$r$ = $\bot \vee \mathcal{S}_{k}[r] = \mathcal{G}[r]$}
\State $r \gets \text{\sc RemoveFromQueue($H$)}$
\EndIf
\State $G' \gets \tupleof{G, \omega, \mathcal{S}, \mathcal{G}}$
\State $P' \gets \text{\sc PushAndPull($G'$, $r$)}$
\State $P \gets P[0:k] + P'[:]$	\Comment Update a set of plans for $\omega$
\EndWhile
\end{algorithmic}
\label{alg:coord}
\end{algorithm}

Alg. \ref{alg:coord} continues until all members in a coupling group $\omega$ reach their final goals. 
The termination condition is checked in line \ref{alg:coord:termination_condition}. 
As long as there exists a robot that has not reached its goal, 
the algorithm will continue with push and pull. 
In Alg. \ref{alg:coord}, $r$ represents the leader of the group $\omega$. 
We remark that there can be cases in which a robot that has already reached its goal may block the path of the leader $r$. 
In this case, push and pull will swap or rotate (similar to the operators in \cite{WildeMW13aamas}) 
robots that have not reached their goals with this blocking robot in order to progress. 
Push and pull also ensures that this blocking robot moves back to its goal afterwards.
}
{
}

In \cite{yu-dars-2014}, we proved that the combination of optimistic and pessimistic decoupling in DisCoF guarantees completeness\footnote{
DisCoF is complete for the class of cooperative pathfinding problems 
in which there are two or more unoccupied vertices in each connected component, 
which is an extension of results in \cite{WildeMW13aamas}.}.


\section{DisCoF$^+$}
\label{sec:tech} 


In this section, we discuss the extensions to DisCoF that are made in a new approach named DisCoF$^+$.
First, we relax the assumption that robots synchronize at every time step (or plan step).
Note that even though robots in different OCs cannot communicate in DisCoF, 
it is assumed that robots act in synchronized time steps 
(i.e., robots are given a fixed amount of time to finish planning and execute a single action at every time step).  
The relaxation of this synchronization is necessary for implementation with real distributed systems,
since we cannot always assume the existence of a global clock and a fixed amount of time for each time step
(e.g., the time required for planning for each robot may be arbitrarily different).
\yhl{We remark that each robot can still access the entire map. We can assume that this information is static such that it is initially given and does not change at all. However, each robot cannot recognize where other robots are if they are out of (indirect) communication and sensing range. This information is dynamic such that it changes arbitrarily.}

Furthermore, we introduce a new decoupling strategy such that robots are also allowed to decouple 
after they form a coupling group (i.e., executing push and pull), 
thus transitioning back to optimistic decoupling from pessimistic decoupling.
This strategy is expected to make DisCoF$^+$ more computationally efficient while achieving higher quality plans that require fewer steps.


\subsection{Asynchronous Time Steps}
\label{sec:asynplanner}
Unlike DisCoF, DisCoF$^+$ allows robots in different OCs to proceed in dependently and asynchronously.
However, robots within the same OC are assumed to still have synchronized plan steps.
This is a reasonable assumption because these robots communicate to coordinate with each other.
As a result of this assumption, robots who finish their current plan step must wait until all others in the OC also finish theirs.
Afterwards, all members of the group start the next plan step at the same time in order to avoid unnecessary conflicts.
We remark that since we assume homogeneous robots, the waiting time at each time step is not significant %
\footnote{\yhl{Heterogeneous robots may have difference in speed, sensing \& communication range and each robot's size, etc. Considering these issues and resolving them are beyond this paper.}}.


\begin{algorithm}[tb]
\caption{DisCoF$^+$ with asynchronous time steps for a robot $i \in \mathcal{R}$, given the environment $G$, its initial location $I$, final destination $F$ and initial plan $P$ from $I$ to $F$; $\gamma$ denotes the contributions values}
\begin{algorithmic}[1]
\State $\tupleof{\psi, \phi, \omega, S[:], \mathcal{G}[:], \gamma[:], k} \gets \tupleof{\emptyset, \emptyset, \emptyset, \emptyset, \emptyset, 0, 0}$
\State $\tupleof{S[i], \mathcal{G}[i]} \gets \tupleof{I, F}$
\State $G' \gets \tupleof{G, \omega, S, \mathcal{G}}$
\State $\tupleof{S, P, \mathcal{W}} \gets \text{\sc ProceedOneStep}(G', P, i, k)$
\While{True}
  \State $\tupleof{\psi, \phi} \gets \text{\sc SenseConflict}(P, i, S, k, \mathcal{W})$ \label{alg:asynplanner:senseconflict}
  \If{$\psi = \emptyset$}
    \State $k \gets k + 1$ \Comment Increase the time step $k$ by $1$
    \State $G' \gets \tupleof{G, \omega, S, \mathcal{G}}$
    \State $\tupleof{S, P, \mathcal{W}} \gets \text{\sc ProceedOneStep}(G', P, i,  k)$
    \State $G' \gets \tupleof{G, \omega, S, \mathcal{G}}$ \Comment Update $G'$ with new $S$
    \State $\tupleof{\gamma, \omega, P} \gets \text{\sc RecomputeCont}(G', P, i, k, \gamma)$
  \Else
    \If{$\omega \neq \emptyset$} \label{alg:asynplanner:involved}
      \State $\omega \gets \omega \cup \phi$ \Comment Merge $\omega$ with OC $\phi$
      \State $G' \gets \tupleof{G, \omega, S, \mathcal{G}}$
      \State $P' \gets \text{\sc PushAndPull}(G', i, \gamma)$
    \Else
      \State $\omega \gets \psi$ \Comment Set $\omega$ to IC $\psi$
      \State $G' \gets \tupleof{G, \omega, S, \mathcal{G}}$
      \State $P' \gets \text{\sc Convergence}(G', i, k, \phi, P, \mathcal{W}, \gamma)$
      \If{$|P'| = 0$} \label{alg:asynplanner:no_local_plan}
        \State $\omega \gets \phi$ \Comment Set $\omega$ to OC $\phi$
        \State $G' \gets \tupleof{G, \omega, S, \mathcal{G}}$
        \State $P' \gets \text{\sc PushAndPull}(G', i, \gamma)$
      \EndIf
    \EndIf
    \State $P \gets P[0:k] + P'[:]$
  \EndIf
\EndWhile
\end{algorithmic}
\label{alg:asyn_planner}
\end{algorithm}

We will explain the difference between DisCoF$^+$ algorithm described in Alg. \ref{alg:asyn_planner} and DisCoF.
In DisCoF$^+$, each robot $i \in \mathcal{R}$ runs the algorithm in Alg. \ref{alg:asyn_planner}.
In line \ref{alg:asynplanner:senseconflict}, 
if there is no conflict sensed in the current location $S[i]$ and local window $\mathcal{W}$ 
(i.e., a fixed region around $S[i]$) of robot $i$, 
such that the IC $\psi$ is empty, robot $i$ can proceed one step ({\sc ProceedOneStep}) forward in its plan $P$.
Afterwards, robot $i$ continues to the next iteration.
On the other hand, 
if a conflict is sensed such that the IC $\psi$ is not empty, 
robot $i$ tries to resolve the conflict after checking if it is already involved in any conflicts at line \ref{alg:asynplanner:involved}.
If it is not involved in any conflict (i.e., it was executing its plan independently),\footnote{
The process when robot $i$ is already involved in a conflict is more involved. Refer to Alg. \ref{alg:asyn_planner} for details.}
it forms a local coupling $\omega$.
It first tries to decouple optimistically through {\sc Convergence}.
If it cannot find a plan $P'$, then it decouples pessimistically through {\sc PushAndPull}.
After finding a plan $P'$, it continues to the next iteration to sense if there are new conflicts.
We remark that our description of {\sc PushAndPull} in Alg. \ref{alg:asyn_planner} is simplified
to show the overall process.
Once {\sc PushAndPull} returns a new plan $P'$ in Alg. \ref{alg:asyn_planner}, 
it contains the individual plan for robot $i$ to move from its location at the time step $k$ to its goal.

\textbf{Correctness:} For Alg. \ref{alg:asyn_planner}, we need to show that whenever there is a conflict, it always returns a valid plan. 
If there is a conflict, in line \ref{alg:asynplanner:involved},
robot $i$ checks if it is already involved in a conflict (with $\omega$). 
If $\omega \neq \emptyset$ (i.e., it is already involved in a conflict), 
we merge the OC (i.e., $\phi$ in Alg. \ref{alg:asyn_planner}) with $\omega$, 
and then call {\sc PushAndPull} for $i$. 
In line \ref{alg:asynplanner:no_local_plan},
if $P'$ is not empty, 
it means that {\sc Convergence} returns a new plan $P'$. 
If $P'$ is empty, then robot $i$ calls {\sc PushAndPull}. 
In both cases, the returned plan $P'$ is either from {\sc Convergence} or {\sc PushAndPull}. 
We have shown that {\sc Convergence} or {\sc PushAndPull} always returns a valid plan in \cite{yu-dars-2014}.

We remark that {\sc ProceedOneStep} always results in the robot proceeding one step forward in its plan.
If robot $i$ has already reached its final goal (while there are robots that still need to reach their goals), 
proceeding one step in this case simply adds a step for robot $i$ to stay.
However, note that when robot $i$ blocks other robots after reaching its goal,
its plan can be updated by these other robots (i.e., forcing robot $i$ to move off its goal temporarily).

\subsection{Communication and Leader Selection}
\label{sec:communication}


There are two major cases in which robots communicate with each other in DisCoF$^+$.
One is to detect predictable conflicts, and another is to synchronize planning and plan execution within the same OC.
Given a robot $i \in \mathcal{R}$, detecting predictable conflicts, performed by {\sc SenseConflict}, requires the following steps:
\begin{enumerate}
\item Check nearby environment (i.e., $\mathcal{W}$) through a sensor for other robots (e.g., a laser sensor);
\item Compute the OC $\phi$ of robot $i$;
\item Communicate with robots in $\phi$ to obtain their plans, then check if predictable conflicts exist among them;
\end{enumerate}


In the above process, 
the first step does not require any communication between robots; it only depends on sensors.
Since robots know the environment (i.e., $G$), they can easily detect when there are moving robots nearby using range sensors.
The second step requires to use the message relay protocol to compute the OC $\phi$.
In the third step, once robot $i$ obtains all the plans of robots in $\phi$, 
it can check these plans against its own plan for predictable conflicts 
(from its current time step to the next $\beta$ steps \cite{yu-dars-2014}).
If conflicts are found with robot $i$'s plan, it forms a IC $\psi$ with the conflicting robots, and then it communicates this back to the robots in the IC $\psi$.
Furthermore, while creating a new plan for robots in IC $\psi$ (i.e., {\sc Convergence}), 
this plan must respect the plans of other robots in the OC $\phi$ of this IC. 
When such a plan cannot be found, 
the set of conflicting robots ($\psi$ initially) is expanded to include other robots in $\phi$
(which are not initially in $\psi$).
\ifthenelse{\boolean{TECHREPORT}}
{
\begin{exmp}[Sensing Conflicts]
Consider the scenario in Fig. \ref{fig:mr:simple}. In this scenario, assume that robot $r_4$ is robot $i$ in the above procedure, so $r_4$ tries to sense a predictable conflict. $r_4$ first senses its nearby environment for other robots. In Fig. \ref{fig:mr:simple}, the local window or the sensing range of $r_4 $ (denoted by $\mathcal{W}$) is shown as the gray region marked with $r_4$, and $r_4$ will detect $r_3$. $r_4$ then computes the OC $\phi$ as $\{r_4, r_3, r_2\}$. Since $r_2$ is not $r_4$'s local window, $r_3$ will relay the communication between $r_2$ and $r_4$. Once $r_4$ obtains both $r_2$ and $r_3$'s plans, it will check their plans against its owns plan for predictable conflicts. In this scenario, $r_4$ will recognize a predictable conflict with $r_3$,
which can be addressed using {\sc Convergence}.
\exmend
\end{exmp}
}
{
}

In the above procedure, the leader who computes the new plan is the robot who first detects the conflict.
Next, the leader tries to resolve the conflict in the IC $\psi$ with {\sc Convergence}.
If it cannot find a new set of plan (i.e., $P'$ in Alg. \ref{alg:asyn_planner}) through {\sc Convergence}, 
it will continue through {\sc PushAndPull} with the OC $\phi$ for the IC $\psi$.
In such cases, we need to choose a new leader (i.e., the robot that moves to its goal first),
which is based on the priorities of subproblems.

The second major case for communication is for synchronized planning and plan execution in an OC. 
This is achieved by {\sc ProceedOneStep}.
Note that robots in different OCs proceed independently and asynchronously.
Since planning and plan execution are synchronized within an OC, 
it guarantees that no collision can occur among robots in the OC.
In {\sc ProceedOneStep}, each robot asks if other robots have finished the execution of their current plan step. 
Once every robot has finished its current plan step, robots can proceed to the next step. 
This requires robots to delay executing the next step until they receive responses from other robots. 
However, 
When robots move out of the communication range, 
they do not synchronize their plan steps anymore. 

\subsection{Flexible Decoupling}
\label{sec:decoupling}

Flexible decoupling is achieved with the help of contribution values.
Contribution values are assigned in DisCoF to each robot in the {\sc Convergence} process 
(in optimistic decoupling),
in which the robots must compute an update to the current plan to avoid potential conflicts.
Contribution values are introduced in DisCoF to ensure that robots are jointly making progress to their goals.
In DisCoF, when the {\sc Convergence} process fails, robots should be in a coupling group, running on the plan computed by {\sc PushAndPull} until they reach their goals.
In DisCoF$^+$, however, robots that are executing {\sc PushAndPull} can again decouple 
by checking whether certain conditions involving the contribution values hold. 


Next, we discuss the new decoupling strategy in DisCoF$^+$,
which is illustrated in the following example.
Suppose that a conflict is predicted between two robots. 
Then, an IC $\psi$ (initially including only the two robots) 
can be formed and there is an associated OC $\phi$ for $\psi$. 
During the {\sc Convergence} process, when the leader of $\psi$ makes a new plan, 
the set of conflicting robots can gradually expand (until becoming $\phi$) 
if the leader cannot find a new plan that avoids the conflict with the current set of conflicting robots,
which is initially $\psi$. 
When a new plan is found, 
DisCoF$^+$ associates each robot with a \textit{contribution value} $\gamma$,
 which captures the individual contribution of the robot to 
 the summation of shortest distances from all robots' current locations to their goal locations.
 
At the very beginning of a problem instance, 
the contribution value $\gamma$ is initialized to be $0$ for all robots.
Given a predictable conflict at time step $k$, 
a set of conflicting robots $\phi$ and a set of current locations $S_k$ for $\phi$,
a set of goal locations $\mathcal{G}$, 
the new plan $\mathcal{Q}$ (where $|\mathcal{Q}| < \beta \in \mathbb{N}$) should satisfy the followings:
\begin{equation}
\sum_{i \in \phi}{\mathcal{C}(S_{k}[i],\mathcal{G}[i]) + \gamma_{k-}[i]} > \sum_{i \in \phi}{\mathcal{C}(S_{k}[i](\mathcal{Q}[i]),\mathcal{G}[i])}
\label{equ:convergence_condition}
\end{equation}
\begin{equation}
\forall i \in \phi, \neg\Delta^i_{k}
\label{equ:not_conflict}
\end{equation}
where $\gamma_{k-}[i]$ is the contribution value that is associated with robot $i$ at the time step $k$, 
$\Delta^i_k$ is a Boolean variable representing whether there is a conflict which is computed based on the updated individual plans and $S_k[i](\mathcal{Q}[i]$) is the local goal for $i \in \phi$.

We remark that while $k$ in Eq. (\ref{equ:not_conflict}) is a constant in DisCoF, in DisCoF$^+$, $k$ represents
the synchronized current time step for the group of robots within $\phi$,
which can be different from different OCs.  
However, note that planning and plan execution are synchronized within $\phi$ until one of robots reaches its goal location. If it reaches its goal location, then it is removed from the group, not being maintained within the communication range anymore.


An interesting point of Eq. \eqref{equ:not_conflict} is that the new plan $\mathcal{Q}$ may not satisfy Eq. (\ref{equ:convergence_condition}) 
during the execution of $\mathcal{Q}$, 
as long as Eq. \eqref{equ:convergence_condition} is satisfied after $\mathcal{Q}$ has completed. 
$\mathcal{Q}$ basically specifies a {local goal} for the robots to reach prior to resuming following their original shortest-path plans again.
Then, the potential conflicts are avoided in the process.
Given a predictable conflict at the current time step and a computed $\mathcal{Q}$, 
the contribution value $\gamma$ while executing the actions in $\mathcal{Q}$ is updated for robot $i$ in $\phi$ as follows:
\begin{equation}
\gamma_{k+\delta}[i] = \mathcal{C}(S_k[i](\mathcal{Q}[i]),\mathcal{G}[i]) - \mathcal{C}(S_{k+\delta}[i],\mathcal{G}[i])
\label{equ:updating}
\end{equation}
where $0 \leq \delta \leq |\mathcal{Q}|$.  
We remark that $\delta$ is a relative time step after the robots have formed an OC. For all robots in a group, $\delta$ is the same.
This update continues until the robot become involved in other conflicts or the value becomes 0.

In DisCoF \cite{yu-dars-2014}, the contribution value $\gamma$ is only used for the {\sc Convergence} process,
and robots do not update their contribution values when a coupling group is formed and robots start {\sc PushAndPull}.
This can lead to inefficient behaviors, e.g., 
when the leader's goal location is located opposite to where the others' goals are located.
\ifthenelse {\boolean{TECHREPORT}}
{
This situation is illustrated in the following example. 
\begin{exmp}[Narrow Corridor]
Figure \ref{fig:mr:narrow_corridor} shows an example of robot $r_2$ in a narrow corridor 
meeting with a coupling group $\{r_1$, $r_3\}$ (executing {\sc PushAndPull}) moving in the opposite direction.
The coupling group $\{r_1$, $r_3\}$ started in the middle corridor, and then $r_1$ became the leader. While $r_1$ pushes $r_3$ to clear away of its path to its goal location $g_1$, it meets $r_2$.
In this case, they will be merged together. 
Suppose that $r_1$ is chosen to be the leader of the new group $\{r_1$, $r_2$, $r_3\}$. 
Until $r_1$ reaches its goal location $g_1$, 
$r_2$ and $r_3$ will be pushed to the end of the middle corridor and then they will be pulled after the intersection $i_1$.
\exmend
\end{exmp}
}
{
}

In DisCoF, the only way to reduce the size of a coupling group is to have the current leader reach its goal.
Then, a new leader will be selected and the remaining robots will follow the new leader to its goal. 
This is clearly an inefficient solution. 
In DisCoF$^+$, we use the contribution values $\gamma$ also in {\sc PushAndPull}, 
such that robots can decouple even before the leader reaches its goal.

\ifthenelse{\boolean{TECHREPORT}}
{
\begin{figure}
\centering
\includegraphics[width=8.5cm]{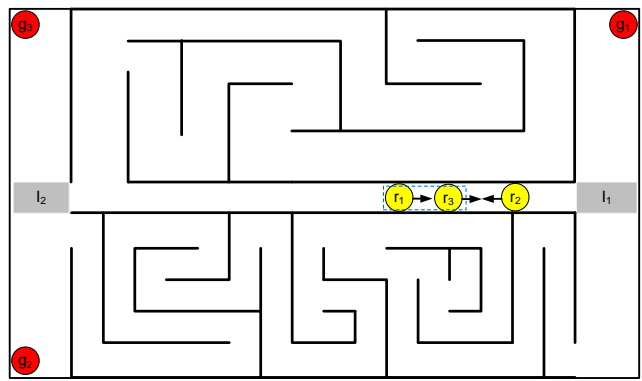}
\caption{Yellow circles are robots and red circles are their goal locations. 
Blue dashed square represents a coupling group of robot $r_1$ and $r_3$. 
This group meets another robot $r_2$ moving in the opposite direction. 
Gray cells represent the intersections of corridors.}
\label{fig:mr:narrow_corridor}
\end{figure}
}
{
}

Next, we discuss how the contribution values can be used in the {\sc PushAndPull} process. 
More specifically, we provide a decoupling condition for a coupling group to check, 
which determines when the robots in the group can decouple while executing the {\sc PushAndPull} process.
Suppose that there is a coupling group $\omega$. 
After $\omega$ computes a new plan $P'$ (in {\sc PushAndPull}), 
each robot in $\omega$ will progress using the plan. 
During this execution, robots continue recomputing their contribution values $\gamma$ as in Eq. \eqref{equ:updating}.
At any step, if the following condition holds, then the group can be decoupled:
\begin{equation}
\sum_{i \in \omega}{\mathcal{C}(S_k[i],\mathcal{G}[i])} + \gamma_{k-}[i] > \sum_{i \in \omega}{\mathcal{C}(S_{k+\delta}[i], \mathcal{G}[i])}
\label{equ:decouple_condition}
\end{equation}
where $k$ is the time step when {\sc PushAndPull} starts planning and $k+\delta$ is the current time step such that $0 < \delta \in \mathbb{N}$.
$\mathcal{C}(S_k[i],\mathcal{G}[i])$ is the length of the shortest-path from location $S_k[i]$ (where conflicts were predicted)
to the goal location $\mathcal{G}[i]$ for each robot $i \in \omega$.
$\gamma_{k-}[i]$ is the contribution value that robot $i \in \omega$ had before the conflicts were predicted.
$\mathcal{C}(S_{k+\delta}[i],\mathcal{G}[i])$ is the length of the shortest-path from the current location $S_{k+\delta}[i]$ to the goal location $\mathcal{G}[i]$ for each robot $i \in \omega$.

Intuitively, Eq. (\ref{equ:decouple_condition}) is the condition 
when the summation of the length of the shortest-path from robots' current locations to their goal locations 
is less than 
the summation of the length of the shortest-path from their original coupling locations to their goal locations 
plus their contribution values just before forming the coupling group.

In Alg. \ref{alg:asyn_planner}, Eq. (\ref{equ:decouple_condition}) is checked inside of {\sc RecomputeCont}. 
If the condition holds, 
then the algorithm returns a new plan $P$ (i.e., a shorted-path plan) with an empty coupling group $\omega$. 
Otherwise, the algorithm returns the current plan $P$ without changing the coupling group $\omega$. 

\ifthenelse {\boolean{TECHREPORT}}
{
\begin{exmp}[Decoupling]
In Fig. \ref{fig:mr:narrow_corridor}, when the coupling group $\{r_1$, $r_3\}$ is merged with $r_2$, 
then conflict locations for $\{r_1, r_2, r_3\}$ and the contribution values (i.e.,  $\gamma$) are saved. 
For a simple illustration, assume that $\gamma = 0$. 
Then, whenever the merged group of robots $\{r_1, r_2, r_3\}$ proceed one time step in their plan 
(which is returned by {\sc PushAndPull}), 
they also check the decoupling condition in Eq.(\ref{equ:decouple_condition}) in {\sc RecomputeCont}. 
However, until the leader $r_1$ reaches its goal location $g_1$, they cannot be decoupled. 
This is because the summation of the distance between robots' locations to their goal locations keeps increasing. 
When $r_1$ reaches its goal location $g_1$, $r_1$ is removed from the group.
Assume that $r_2$ is elected as a new leader of the group.
Then, $r_3$ will be pulled until they reach the conflict location where they met previously. (See the place where they are placed in the Fig. \ref{fig:mr:narrow_corridor})
After passing the conflict location, 
$r_3$ and $r_2$ can be decoupled since 
Eq. \eqref{equ:decouple_condition} holds. 
Consequently, from the intersection $I_2$, 
$r_2$ and $r_3$ can move independently to their goal locations. 
\exmend
\end{exmp}
}
{
}

When a coupling group is decoupled and it immediately predicts a conflict in the next iteration, 
it uses the conflict resolution process through {\sc Convergence}, just as when fully decoupled robots have predicted conflicts.
Even though we discussed the correctness of DisCoF$^+$ (Alg. \ref{alg:asyn_planner}), 
we also need to show that this new decoupling strategy is not subject to live-locks 
(i.e., robots are always making joint progress to the goals).

\begin{theorem}
The decoupling condition in Eq.\eqref{equ:decouple_condition} ensures that robots in the group gradually progress to their final goals.
\end{theorem}


\ifthenelse {\boolean{TECHREPORT}}
{
\begin{proof}
From Eq. \eqref{equ:convergence_condition} and Eq. \eqref{equ:updating}, 
we know that each robot in the group gradually moves towards its final goal. 
Here, we show that Eq. \eqref{equ:decouple_condition} does not prevent any group member from reaching its goal. 
Given that we use the contribution value $\gamma_{k-}$ when a coupling group is formed, in order to satisfy Eq. (\ref{equ:convergence_condition}) when decoupling, either robots can all execute their original plans or {\sc Convergence} must return a new plan which progresses robots to their local goals.
First, their original plans definitelly make progress.
Second, consider the case when it takes the new plan from {\sc Convergence}.
After progressing through the new plan, all the robots in the group will reach their local goals.
Then, the summation of the distance from their current locations (which are their local goals) to their final goals is smaller than the summation of the distance from their locations (where they predicted the conflicts) to their final goals plus their contribution values $\gamma$ before forming the coupling group.
Hence, we can conclude that robots would be making joint progress to their goals. 
Hence, the decoupling condition Eq. \eqref{equ:decouple_condition} 
does not prevent the group members from progressing to their final goals.
\end{proof}
}
{
\begin{proof}
For detail, see \cite{Kim2015Arxiv}.
\end{proof}
}
\section{RESULTS}
\label{sec:results}

In this section, we will show experimental results. First, we will show a simulation result. Second, we will provide a result of numerical experiments.

\subsection{Simulation Result}

\begin{figure}
\centering
\includegraphics[width=8cm]{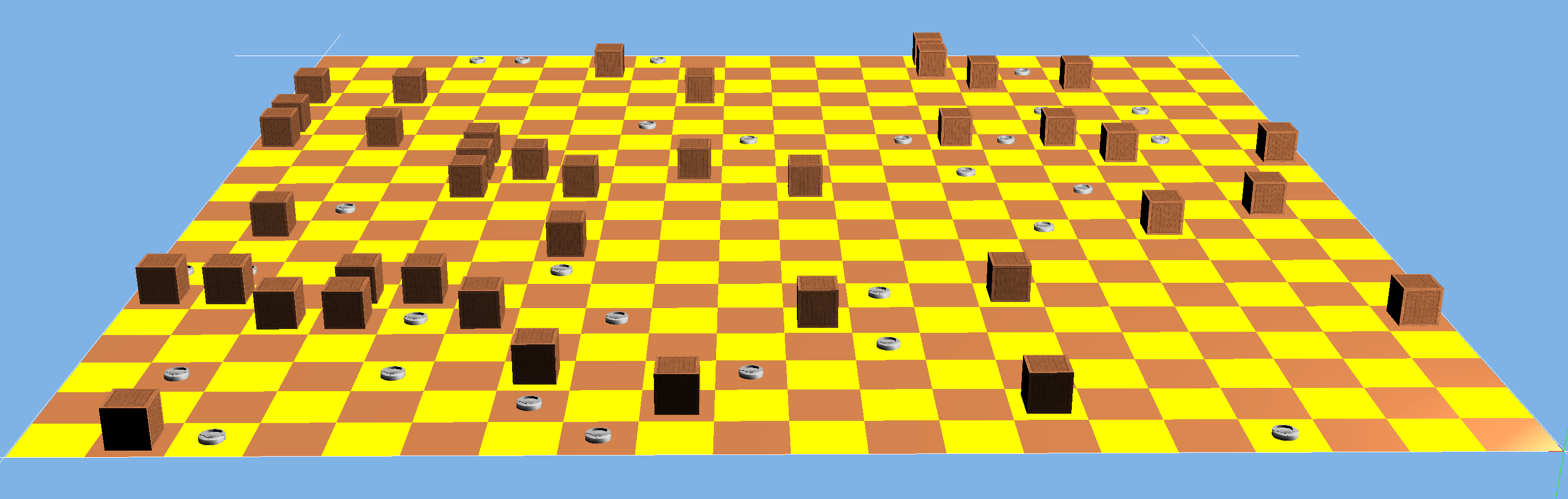}
\caption{Simulation environment represents a $20 \times 20$ grid world with 10\% wooden box obstacles. In this environment, there are 30 iRobot Create finding their path to their goal positions.}
\label{fig:mr:simulator_screenshot}
\end{figure}

The simulation shown in Fig. \ref{fig:mr:simulator_screenshot} was created using Webots 7.3.0 and the included iRobot Create models. A grid environment was modelled which contained 30 iRobots and 40 obstacles placed at random (but solvable) locations. Each iRobot was running with a controller which implemented DisCoF+, however, one exception was made. Rather than being completely distributed and simulating ad hoc networks and localization, the robots communicated with a central supervisor which provided this information as well as synchronization for robots involved in a conflict. Robots in different outer closures acted completely asynchronously, but robots in the same outer closure were synchronized if a conflict was detected between any of the member robots. Ultimately, this concession will be replaced with simulated sensors and a fully distributed environment, but for now it still provides valuable results.

The target computer for the simulation was a MacBook Pro running Mac OS X 10.10.2 with a 2.3GHz i7 and 16GB of RAM. The simulation was run two times: once with decoupling enabled and once with decoupling disabled. Decoupling enabled yielded a total simulation duration of 3 minutes and 23 seconds. Out of all robots, the maximum number of steps required to reach its destination was 40. Decoupling disabled yielded a total simulation duration of 5 minutes and 1 second. Out of all robots, the maximum number of steps required to reach its destination was 54.

These results are interesting for two reasons: one is that decoupling enabled performs significantly better, another is the ratio of decoupling enabled vs decoupling disabled simulation time compared to that of maximum steps. With decoupling the simulation took 67\% of the time and 74\% of the steps that decoupling disabled did. The reason for this discrepancy is that with decoupling enabled there are more “stay” actions in which a robot’s action is to stay where it is at. Since robots are asynchronous except for when they are in a conflict, this means robots will take less time to complete a plan with stay actions compared to one that doesn't. It is expected that environments requiring more complex plans will benefit from this fact even more.


%
We provide the demo video for this simulation. Refer the videos to the following url: \url{https://www.assembla.com/spaces/discof/wiki/DisCoF_Plus}.

\subsection{Numerical Experiments}

\begin{table*}
\begin{center}
\resizebox{\textwidth}{!}{%
\begin{tabular}{|c|cc|cc|cc|cc|cc|cc|}
\hline
		& \multicolumn{6}{c|} {DisCoF}	& \multicolumn{6}{c|} {DisCoF$^+$ (DisCoF$^+$/DisCoF)} \\
\hline
		& \multicolumn{2}{c|} {COMP. TIME}	& \multicolumn{2}{c|} {STEPS}  & \multicolumn{2} {c|} {APPROX. RUN TIME} & \multicolumn{2}{c|} {COMP. TIME} & \multicolumn{2}{c|} {STEPS} & \multicolumn{2} {c|} {APPROX. RUN TIME} \\
\hline
OBSTACLES	& AVG		& STD	 	& AVG		& STD		& AVG		& STD		& AVG				& STD				& AVG				& STD				& AVG				& STD	\\
\hline
5\%			& 10.064 	& 8.405		& 352.35	& 356.207	& 1771.815	& 1788.861 	& 10.733 (1.0086)	& 22.068 (0.931)	& 63.95	(0.4266) 	& 80.632 (0.356)	& 330.483 (0.43)	& 423.885 (0.3555)	\\
\hline
10\% 		& 13.19		& 10.372	& 521.1		& 521.24 	& 2618.69	& 2615.82	& 14.37	(1.061)		& 36.52	(1.538)		& 73.51	(0.344)		& 108.93 (0.346) 	& 381.92 (0.348)	& 579.065 (0.348)	\\
\hline
15\%		& 17.6318 	& 13.296 	& 653.67  	& 580.01 	& 3285.982	& 2911.463	& 23.92 (1.217)		& 49.768 (1.3)		& 99.18 (0.294) 	& 157.356 (0.312)	& 519.82 (0.3)		& 831.07 (0.314)	\\
\hline
20\%		& 26.39		& 14.009	& 954.46	& 620.08 	& 4798.691	& 3111.208	& 52.391 (1.942)	& 75.8 (2.3989)		& 175.9192 (0.2427)	& 218.859 (0.3132)	& 931.987 (0.2535)	& 1161.61 (0.3242)	\\
\hline
\end{tabular}
}
\end{center}
\caption{Numerical Experiments:
Comp. Time represents the total computation time in second, Steps represent concurrent time steps for entire robots' plan, and Approx. Run Time represents approximate running time in second. For Comp. Time, Steps and Approx. Run Time, DisCoF and DisCoF$^+$ have average and standard deviation. 
}
\label{table:numerical_expriments2}
\end{table*}

In order to evaluate the improvement of DisCoF$^+$ over DisCoF, we execute a number of numerical experiments.
For these experiments, we used a Core i7 CPU 3.2 Ghz with 8GB memory and 240 GB SSD in Cygwin environment which runs on Windows 8.1. Our prototype implementation is written in Python 2.7.2.


Since we only want to get the total concurrent steps and the computation time for these experiments, instead of using Webots simulator, we used a simple discrete time simulator which does not simulate the phisics of the robots or any communication between the robots. Hence, we are comparing the total number of steps and the computation times between DisCoF and DisCoF$^+$.
%

As a result of this implementation, we compute an approximate running time by assuming that each robot move takes 5 seconds. Then, for each problem instance, the running time is equal to the sum of the computation time and the maximum number of moving steps for a robot times 5.

We show that the decoupling approach improves upon the previous approach (DisCoF) in. In order to perform the experimental analysis, instead of scaling up the number of robots, we increase the density of the environment. That is, we increase obstacle rates in the environment. Given 30 robots and 20 x 20 grid environment with their goal locations and obstacles, the Table \ref{table:numerical_expriments2} shows the result. The obstacle rates changed from 5\% to 20\%. For this test, we randomly generated 100 instances for each obstacle rate. Obstacle locations were also randomly chosen.

The time ratio in Table \ref{table:numerical_expriments2} indicates that if the environment is less populated, then decoupling makes better quality plans in terms of the total number of concurrent steps and the total computation time of plans.

However, this result also shows an interesting property of DisCoF$^+$.
When the environment gets denser, the decoupling method does not always reduce the total computation times. This is because in dense environments groups that decouple may have to recouple with a higher frequency. When it is re-coupled, a group should make a new plan which requires extra computation time.
On average, DisCoF$^+$ needed 34.4\% steps less than DisCoF's result.

\section{CONCLUSIONS}
In this paper, we introduced DisCoF$^+$ which is an asynchronous extension of our previous work. We also introduced a strategy of decoupling in DisCoF$^+$. Through simulations, we showed how DisCoF$^+$ works in a simulated grid environment to resolve predictable conflicts in a distributed fashion. We also provided numerical experiments to compare DisCoF with DisCoF$^+$. In moderately populated environments, the decoupling approach shows bettered results than DisCoF. In future work, we plan to devise different approaches for the decoupling such as more strict decoupling conditions and also a heuristic for ordering robots while performing {\sc PushAndPull} so that when at any point time a decoupling occurs, the conflicts are minimized.
\label{sec:conclusion}

\section*{Acknowledgments}
The authors would like to thank the anonymous reviewers for their detailed comments and suggestions.

\bibliographystyle{abbrv}
\bibliography{bibliography}

\begin{thebibliography}{10}

\bibitem{AyanianRK12wdecns}
N.~Ayanian, D.~Rus, and V.~Kumar.
\newblock Decentralized multirobot control in partially known environments with
  dynamic task reassignment.
\newblock In {\em 3rd IFAC Workshop on Distributed Estimation and Control in
  Networked Systems}, 2012.

\bibitem{bnaya}
Z.~Bnaya and A.~Felner.
\newblock Conflict-oriented windowed hierarchical cooperative {A}$^*$.
\newblock In {\em Proceedings of the 2014 IEEE International Conference on
  Robotics and Automation}, 2014.

\bibitem{clark2003}
C.~Clark, S.~Rock, and J.-C. Latombe.
\newblock Motion planning for multiple mobile robots using dynamic networks.
\newblock In {\em Proceedings of the IEEE International Conference on Robotics
  and Automation}, volume~3, pages 4222--4227, Sep. 2003.

\bibitem{WildeMW13aamas}
B.~de~Wilde, A.~W. ter Mors, and C.~Witteveen.
\newblock Push and rotate: Cooperative multi-agent path planning.
\newblock In {\em 12th International Conference on Autonomous Agents and
  Multiagent Systems}, 2013.

\bibitem{DesarajuH12ar}
V.~R. Desaraju and J.~P. How.
\newblock Decentralized path planning for multi-agent teams with complex
  constraints.
\newblock {\em Autonomous Robots}, 32(4):385--403, 2012.

\bibitem{hopcroft1984}
J.~Hopcroft, J.~Schwartz, and M.~Sharir.
\newblock On the complexity of motion planning for multiple independent
  objects; pspace- hardness of the "warehouseman's problem".
\newblock {\em The International Journal of Robotics Research}, 3(4):76--88,
  1984.

\bibitem{jansen2008}
R.~Jansen and N.~Sturtevant.
\newblock A new approach to cooperative pathfinding.
\newblock In {\em Proceedings of the 7th International Joint Conference on
  Autonomous Agents and Multiagent Systems}, AAMAS, pages 1401--1404, Richland,
  SC, 2008. International Foundation for Autonomous Agents and Multiagent
  Systems.

\bibitem{LiuS13ijrr}
L.~Liu and D.~A. Shell.
\newblock Physically routing robots in a multi-robot network: Flexibility
  through a three-dimensional matching graph.
\newblock {\em The International Journal of Robotics Research},
  32(12):1475--1494, 2013.

\bibitem{LunaB11iros}
R.~Luna and K.~Bekris.
\newblock Efficient and complete centralized multirobot path planning.
\newblock In {\em IEEE/RSJ Int. Conf. on Intelligent Robots and Systems}, 2011.

\bibitem{frazzoli}
M.~Otte, J.~Bialkowski, and E.~Frazzoli.
\newblock Any-com collision checking: Sharing certificates in decentralized
  multi-robot teams.
\newblock In {\em Proceedings of the 2014 IEEE International Conference on
  Robotics and Automation}, 2014.

\bibitem{Parker09}
L.~E. Parker.
\newblock {\em Encyclopedia of Complexity and System Science}, chapter Path
  Planning and Motion Coordination in Multiple Mobile Robot Teams.
\newblock Springer, 2009.

\bibitem{peasgood2008}
M.~Peasgood, C.~Clark, and J.~McPhee.
\newblock A complete and scalable strategy for coordinating multiple robots
  within roadmaps.
\newblock {\em IEEE Transactions on Robotics}, 24(2):283--292, April 2008.

\bibitem{ryan2007}
M.~Ryan.
\newblock Graph decomposition for efficient multi-robot path planning.
\newblock In {\em Proceedings of the 20th International Joint Conference on
  Artifical Intelligence}, pages 2003--2008, San Francisco, CA, USA, 2007.
  Morgan Kaufmann Publishers Inc.

\bibitem{Sharon201540}
G.~Sharon, R.~Stern, A.~Felner, and N.~R. Sturtevant.
\newblock Conflict-based search for optimal multi-agent pathfinding.
\newblock {\em Artificial Intelligence}, 219(0):40 -- 66, 2015.

\bibitem{Silver05aiide}
D.~Silver.
\newblock Cooperative pathfinding.
\newblock In {\em Conference on Artificial Intelligence and Interactive Digital
  Entertainment}, 2005.

\bibitem{standley2010}
T.~Standley.
\newblock Finding optimal solutions to cooperative pathfinding problems.
\newblock In {\em AAAI Conference on Artificial Intelligence}, 2010.

\bibitem{standley2011}
T.~Standley and R.~Korf.
\newblock Complete algorithms for cooperative pathfinding problems.
\newblock In {\em Proceedings of the 22nd International Joint Conference on
  Artifical Intelligence}, 2011.

\bibitem{sturtevant2006}
N.~Sturtevant and M.~Buro.
\newblock Improving collaborative pathfinding using map abstraction.
\newblock In {\em Artificial Intelligence and Interactive Digital Entertainment
  (AIIDE)}, pages 80--85, 2006.

\bibitem{wang08}
K.~C. Wang and A.~Botea.
\newblock Fast and memory-efficient multi-agent pathfinding.
\newblock In {\em International Conference on Automated Planning and
  Scheduling}, pages 380--387, 2008.

\bibitem{YuL12wafr}
J.~Yu and S.~M. LaValle.
\newblock Multi-agent path planning and network flow.
\newblock In {\em Algorithmic Foundations of Robotics X}, volume~86, pages
  157--173. Springer, 2013.

\bibitem{yu-dars-2014}
Y.~Zhang, K.~Kim, and G.~Fainekos.
\newblock Discof: Cooperative pathfinding in distributed systems with limited
  sensing and communication range.
\newblock In {\em to appear in International Symposium on Distributed
  Autonomous Robotic Systems}, 2014.

\end{thebibliography}

\end{document}